\documentclass[10pt,twocolumn,letterpaper]{article}

\usepackage{cvpr}
\usepackage{times}
\usepackage{epsfig}
\usepackage{graphicx}
\usepackage{amsmath}
\usepackage{amssymb}
\usepackage{subfigure}
\usepackage{booktabs}
\usepackage{multirow}

\usepackage[breaklinks=true,bookmarks=false]{hyperref}

\cvprfinalcopy

\begin{document}

\title{Boundary-Aware Salient Object Detection via Recurrent Two-Stream Guided Refinement Network}

\author{Fangting Lin, Chao Yang, Huizhou Li, Bin Jiang\\
College of Computer Science and Electronic Engineering, Hunan University\\
{\tt\small linfangting@hnu.edu.cn, yangchaoedu@hnu.edu.cn, lihz@hnu.edu.cn, jiangbin@hnu.edu.cn}
}

\maketitle

\begin{abstract}
Recent deep learning based salient object detection methods which utilize both saliency and boundary features have achieved remarkable performance. However, most of them ignore the complementarity between saliency features and boundary features, thus get worse predictions in scenes with low contrast between foreground and background. To address this issue, we propose a novel Recurrent Two-Stream Guided Refinement Network (RTGRNet) that consists of iterating Two-Stream Guided Refinement Modules (TGRMs). TGRM consists of a Guide Block and two feature streams: saliency and boundary, the Guide Block utilizes the refined features after previous TGRM to further improve the performance of two feature streams in current TGRM. Meanwhile, the low-level integrated features are also utilized as a reference to get better details. Finally, we progressively refine these features by recurrently stacking more TGRMs. Extensive experiments on six public datasets show that our proposed RTGRNet achieves the state-of-the-art performance in salient object detection.
\end{abstract}

\section{Introduction}
Salient object detection~\cite{zhao2019pyramid,fan2019shifting,wang2017video} aims to find one or more objects which attract most attention in a given image or video and then segment these objects out. The core challenge of this task is how to keep a high accuracy when detecting salient objects in different complex scenes (e.g. have a low contrast with the background or in cluttered scenes). Therefore, it is necessary to design an effective network to achieve accurate prediction.

Traditional methods mostly use hand-crafted features (e.g. colors, intensity, orientations, or others) to distinguish the difference between background and salient objects. However, these hand-crafted features lack  high-level semantic information and usually fail to detect salient objects with complex scenes. Fortunately, along with the advancement of deep learning approaches, fully convolutional neural networks (FCNs)~\cite{long2015fully} have shown impressive results in many computer vision tasks, e.g. pose estimation~\cite{cao2017realtime} and image inpainting~\cite{liu2019coherent}. Motivated by these achievements, FCN-based methods are widely applied for salient object detection in recent years. Nevertheless, FCN-based methods mainly focus on the design of the fusion mechanism of multi-level features, but ignore that the boundary information is difficult to capture. The lack of boundary information then results in the poor boundary segmentation results. 

Recently, there has been increasing interst in jointly refining salient object and boundary. Some methods consider both salient object features and boundary features to improve the segmentation results: adding extra boundary detection branch~\cite{wang2019focal,su2019selectivity}, designing boundary aware loss~\cite{Qin_2019_CVPR}, introducing the external edge labels~\cite{Liu2019PoolSal,WuRunMin_2019_CVPR}, and considering the complementarity between the salient object features and boundary features~\cite{zhao2019EGNet}. Although these single-stage methods can effectively improve the segmentation result, only using boundary features to refine salient object features is still not enough to enhance the prediction. On the other hand, \cite{wang2018salient,deng18r} adopt recurrent mechanism to  progressively refine the saliency map. Although these multi-stage methods can effectively improve the prediction results, the pixel around boundary are still hard to capture. Besides, \cite{Wu_2019_ICCV} proposes a recurrent network that mutually complement the salient object information and boundary information at each recurrent step. Although this method achieves the state-of-the-art performance, it still suffer from coarse edge in complex scenes, due to its lack of suitable guide of low-level features.

\begin{figure*}
\begin{center}
\includegraphics[width=1\textwidth]{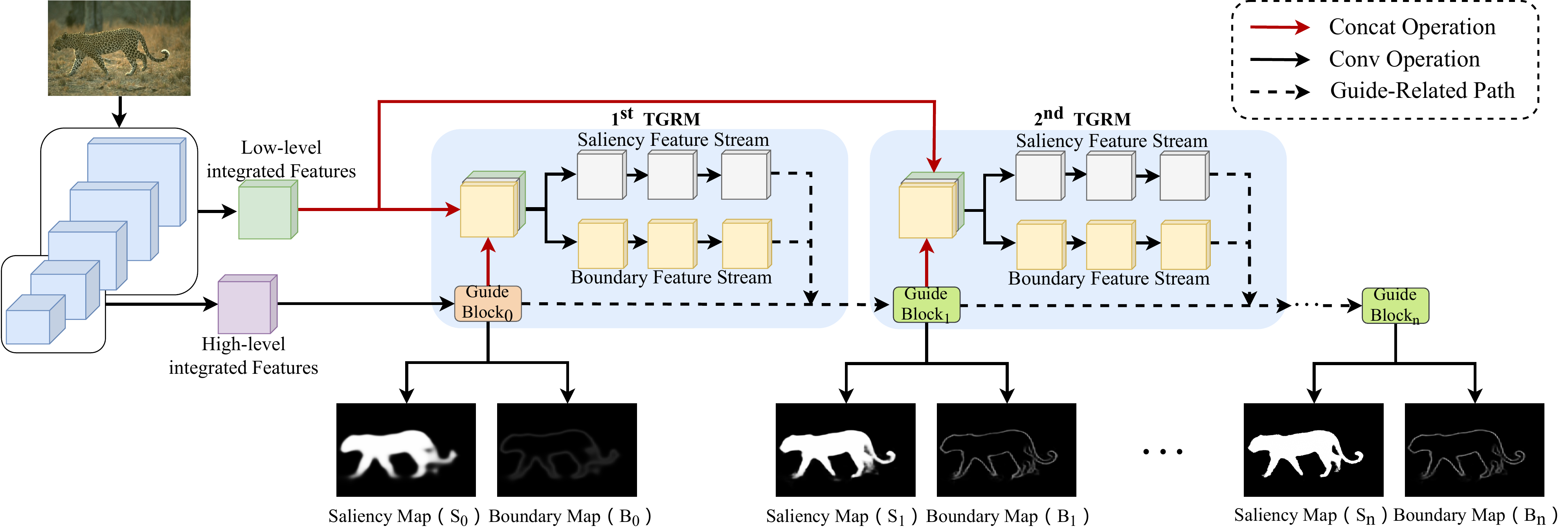} 
\end{center}
   \caption{The overall pipeline of our proposed Recurrent Two-Stream Guided Refinement Network: RTGRNet. We first extract multi-level features from feature extraction network and merge these features together to form the low-level and high-level integrated features. The initial saliency and boundary features are generated by high-level integrated features. Then we utilize stacked TGRMs to refine these features by using difference feature streams. In each TGRM, low-level integrated features and Guide Block producted by previous TGRM are used to guiding network to consider global refinement and boundary enhancement simultaneously. We finally generate the saliency and boundary maps at each recurrent step, and use the saliency map producted by the last recurrent step as our final result.}
\label{fig:short}
\end{figure*}

Motivated by the above observations, in this paper, we propose a novel Recurrent Two-Stream Guided Refinement Network (RTGRNet) that consists of Two-Stream Guided Refinement Modules (TGRMs). RTGRNet implements salient object detection by iterating TGRMs which simultaneously model the complementary information of saliency and boundary in a single network. First, inspired by \cite{deng18r}, as the low-level integrated features can capture more saliency details and the high-level integrated features can reduce non-salient regions, we extract low-level integrated features and high-level intergrated features as the input of our network. Then, the sequences of TGRMs progressively refine these features in two streams: saliency feature stream and boundary feature stream. Each TGRM consists of a Guide Block, and two feature streams of boundary and salient object. The first Guide Block is initialized by the saliency features and boundary features from the high-level integrated features, then we use the Guide Block to refine the low-level integrated features. In both saliency and boundary feature streams, the refined low-level integrated features and the feature after the Guide Block will be concatenated and fed to the next Guide Block. With this recurrently stacking stategy, the saliency features and boundary features will be progressively refined. Meanwhile, each Guide Block will be optimized by the saliency maps and boundary maps. At last, the Guide Block of the last TGRM is taken as the final output.

Our contributions of this work are mainly summarized as follows:
\begin {itemize}
	\item We propose a novel module namely Two-Stream Guided Refinement Module (TGRM), which simultaneously models the  complementary information of boundary and salient object to jointly guide the prediction of different tasks. In the TGRM, we design a Guide Block and two feature streams of saliency and boundary to simultaneously refine the salient object and boundary features.
	\item We design a Recurrent Two-Stream Guided Refinement Network: RTGRNet, which consists of a sequence of TGRMs to progressively refine both the salient object features and boundary features. The low-level integrated features are utilized at each recurrent step as a reference to get better details.
	\item Experiments on six public datasets show that our proposed RTGRNet achieves the state-of-the-art results in terms of both F-measure and MAE evaluation measures. As a side contribution, we will release our code.
\end {itemize}
\begin{figure*}
\begin{center}
\subfigure[RRB]{
\label{figa} 
\includegraphics[width=0.32\textwidth]{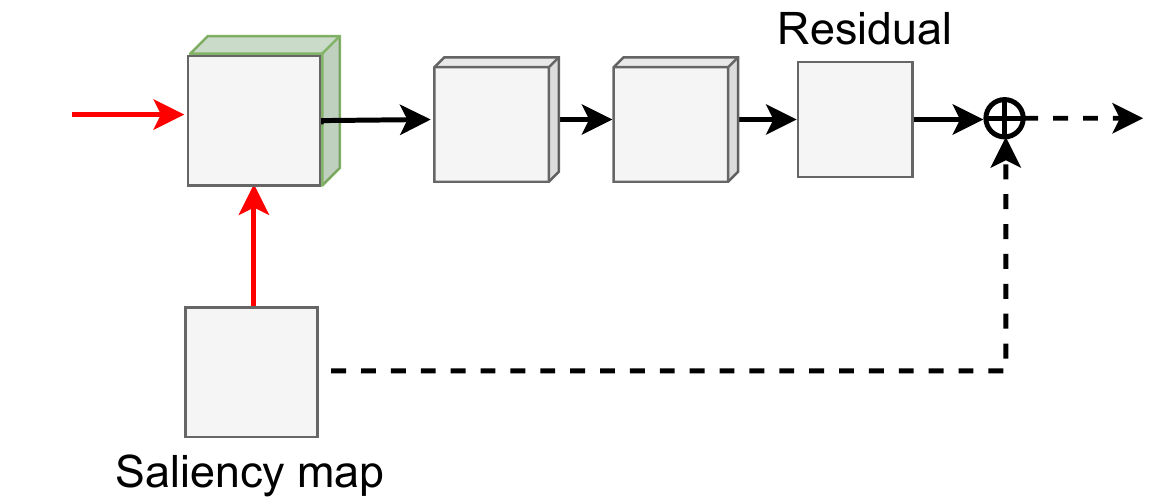}}
\subfigure[SGRM]{
\label{figb} 
\includegraphics[width=0.32\textwidth]{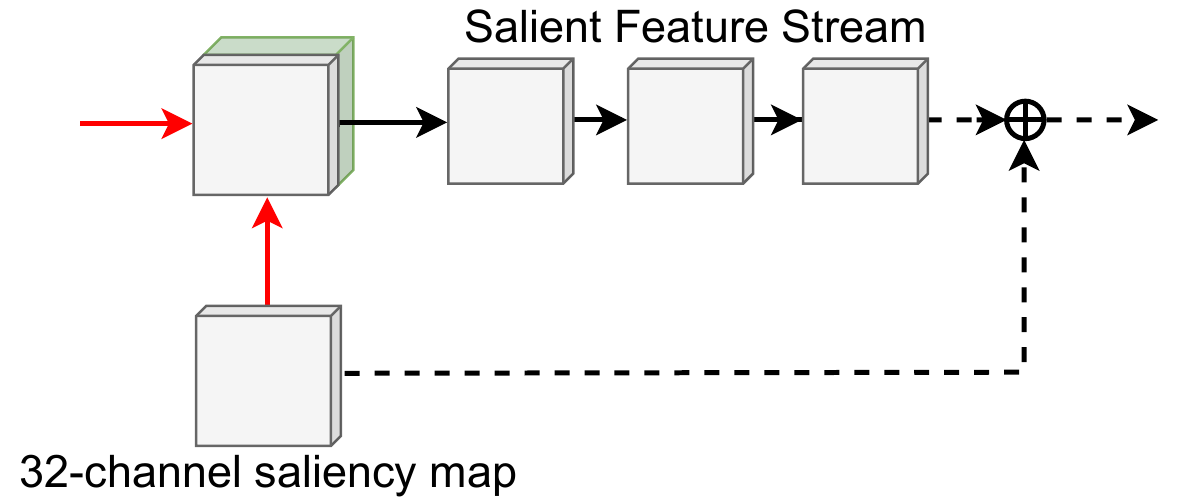}}
\subfigure[TGRM]{
\label{figc} 
\includegraphics[width=0.32\textwidth]{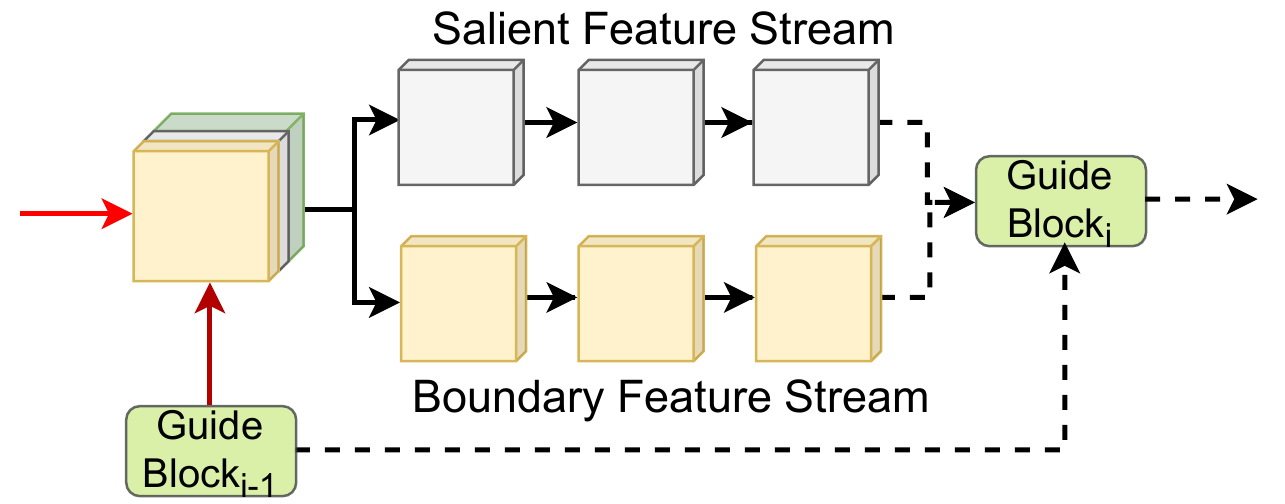}}
\end{center}
   \caption{Detailed illustration of different recurrent refinement modules : (a) Residual Refinement Block (RRB) (b) Single-Stream Guided Refinement Module(SGRM) (c) Two-Stream Guided Refinement Module(TGRM). SGRM and TGRM use low-level intergrated feature for refinement while RRB uses low-level integrated feature and high-level integrated feature to alternatively refine the initial saliency map. Saliency feature stream and boundary feature stream in SGRM and TGRM share the same design.}
\label{fig:short}
\end{figure*}
\section{Related work}
Since Itti $et$ $al.$~\cite{itti1998model} proposed the first salient object detection method, many hand-crafted feature based traditional methods~\cite{wang2011image,cheng2014global} have been proposed in the past two decades. While these methods suffer from inflexible feature design, thus limit their ability in capturing salient objects in different scenes. Boorji $et$ $al.$ provide a comprehensive survey in ~\cite{borji2014salient}. As recent deep learning based salient object detection methods have achieved the state-of-the-art performance, then in the following, we focus on these deep salient object detection methods.

{\bf FCN-based Methods:} As Fully Convolutional Networks (FCNs) are able to capture richer spatial and multi-scale information, they achieve better results in salient object detection. Early FCN-based methods \cite{wang2016saliency,zhang2017deep} detecte salient objects by combining traditional saliency prior information to achieve better prediction than other traditional methods. Meanwhile, some works attempt to fuse features at different levels, in order to design a fast and high-precision salient object detection network. Hou $et$ $al.$~\cite{HouPami19Dss} propose a strongly supervised short-connection structure on the basis of HED~\cite{xie15hed}, which enables the model to effectively utilize high-level semantic features and low-level detail features. Liu $et$ $al.$~\cite{liu2016dhsnet} propose an end-to-end salient object detection method, which first locates the salient object globally, and then progressively refines the details of salient object by integrating the local context. However, these methods directly combine features of different levels, then noise is easy to be introduced into the model. These noises may causes some wrong prediction when positioning and segmenting salient objects.

{\bf Deep Recurrent Methods:} Some methods have been proposed in recent year by using recurrent refinement mechanism to reduce unnecessary information and to progressively refine the saliency map. Wang $et$ $al.$~\cite{wang2017stagewise} propose a multi-stage recurrent refinement network that uses low-level features to refine the initial saliency map, and achieves better accuracy of salient object detection. Deng $et$ $al.$~\cite{deng18r} design a new residual refinement block (RRB) to learn the residual between the ground truth and the saliency map in each recurrent step. Besides, a novel recurrent residual refinement network (R$^3$Net) is proposed to progressively refine the saliency map by constructing a sequence of RRBs to alternatively use the low-level features and high-level features, which can more accurately detect salient regions.

{\bf Deep Boundary-Aware Methods:} More recently, some researchers also attemp to use boundary information to achieve better representation of segmentation feature. Wang $et$ $al.$~\cite{wang2019focal} propose a boundary guided network which jointly learn two sub-network for salient objects and boundaries. Qin $et$ $al.$~\cite{Qin_2019_CVPR} propose a Coarse-to-Fine Network (BASNet) which designs a novel hybrid loss to enhance the prediction near the boundary by a larger loss. Liu $et$ $al.$~\cite{Liu2019PoolSal} predict the boundary maps by a external edge labels and then incorporate it with U-Net architecture to detect salient objects. Zhao $et$ $al.$~\cite{zhao2019EGNet} propose an edge guidance network which considers the complementarity between salient objects and boundary. Wu $et$ $al.$~\cite{Wu_2019_ICCV} propose an stacked cross refinement network to mutually complement the information of salient object and boundary while recurrently refine these information.
\section{RTGRNet}
In this section, we first describe the architecture overview of our proposed RTGRNet, which aims to progressively refine the initial saliency map and enhance the prediction of boundary simultaneously. The detail of our Two-Stream Guided Refinement Module in Sec. 3.2 and Recurrent Refinement Network in Sec. 3.3. The formulation of our loss is presented in Sec. 3.4.
\subsection{Overview of Network Architecture}
The architecture of our proposed RTGRNet is illustrated in Fig 1. The RGNRNet mainly consists of recurrent TGRMs, trying to progressively refine saliency and boundary features. Meanwhile, the low-level integrated features are utilized at each recurrent step as a reference to get better details. Each TGRM exploits a Guide Block and a two-stream structure to predict the saliency map and the boundary map. Specifically, the Guide Block is to generate the guide features of salient object and boundary, so as to refine the low-level features. The two-stream structure are designed to mutually complement the saliency features and boundary features.
\subsection{Two-Stream Guided Refinement Module (TGRM)}
In the literature, ~\cite{deng18r} proposes a Residual Refinement Block (RRB) to progressively refine the saliency map (see Fig,2 a). However, the learned residual, defined as the difference between the ground truth and the previous saliency map is used to guide the subsequent recurrent step, which is compressed into one channel before adding the previous saliency map to get the next saliency map. This operation will cause a heavily information loss.

To solve this problem, we first propose a Single-Stream Guided Refinement Module (SGRM) as shown in Fig. 2 b. In order to reduce the information loss of the model, we add a layer of convolution before we generate the saliency map and the residual, which outputs a 32-channels salient features. These wider saliency map can retain more meaningful information for providing better guidance in the recurren step. Finally, we employ a convolution layer to generate saliency map and boundary map at each recurrent step.

After that, We develop a TGRM at each recurrent step to refine both saliency and boundary features, and adopt two types of features to mutually enhance the prediction of another tasks, as shown in Fig. 2 c. TGRM consists of a Guide Block and two feature streams of saliency and boundary. The Guide Block is designd to complement missing information for two different feature streams. Each feature stream outputs the refined low-level saliency and boundary features. To explain TGRM more clearly, we will first introduce the internal structure of the Guide Block (see fig. 3). Formally, a Guide Block is defined as:

\begin{figure}
\begin{center}
\subfigure[i=0]{
\label{figa_gb} 
\includegraphics[width=0.20\textwidth]{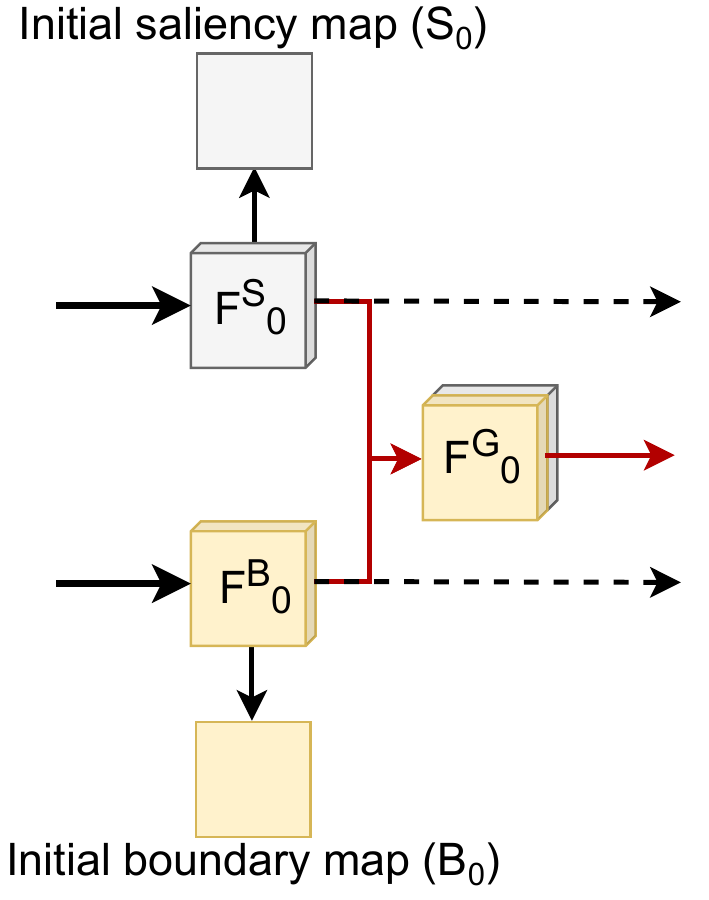}}
\subfigure[$i\not=0$]{
\label{figb} 
\includegraphics[width=0.25\textwidth]{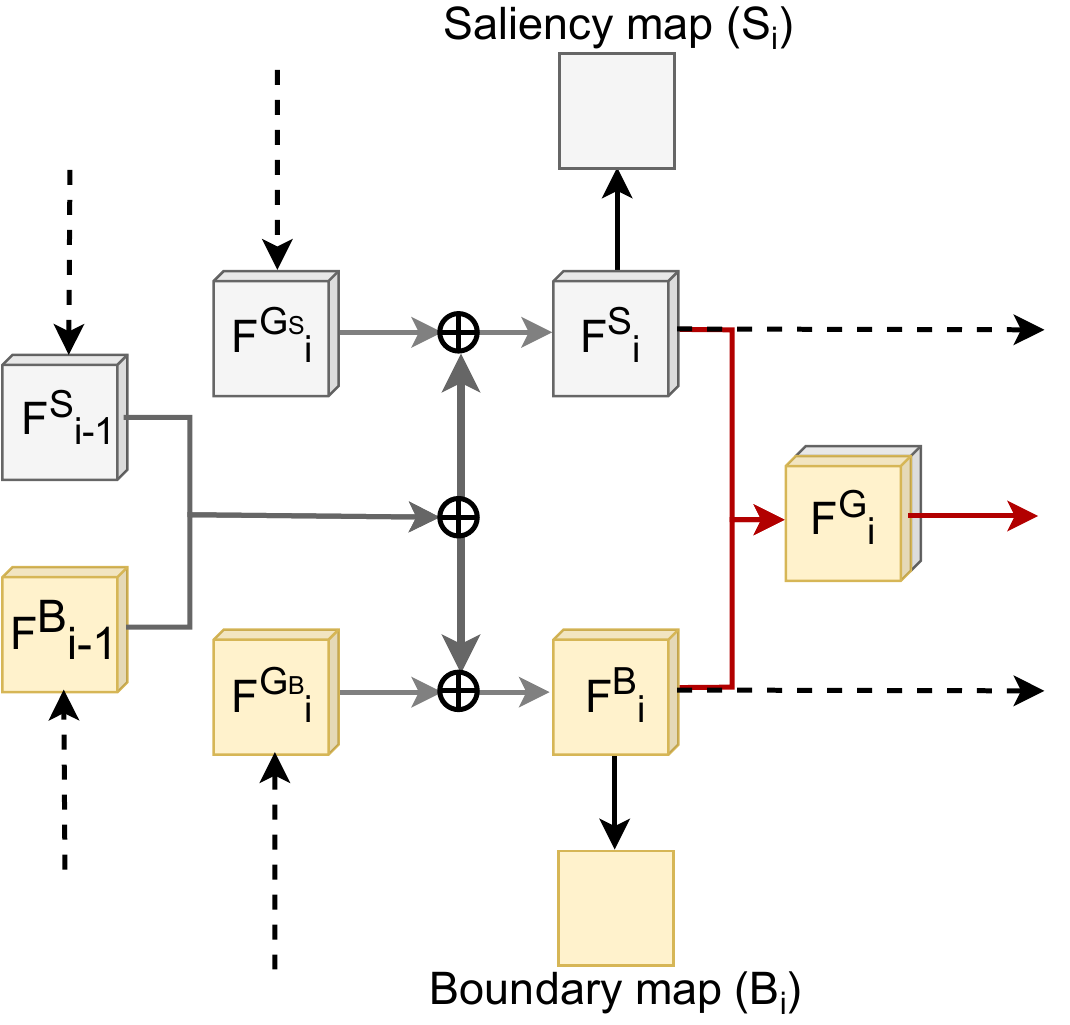}}
\end{center}
   \caption{Detailed illustration of Guide Block. Each Guide Block outputs three guide features, Saliency map and Boundary map at i-th recurrent step.}
\end{figure}
\begin{gather}
{\mathop{{{\mathop{{F}}\nolimits^{{S}}}}}\nolimits_{{i}}={ \left\{ \begin{array}{*{20}{l}}
{\mathop{{{\mathop{{F}}\nolimits^{{S}}}}}\nolimits_{{i-1}} \oplus \mathop{{{\mathop{{F}}\nolimits^{{B}}}}}\nolimits_{{i-1}} \oplus \mathop{{{\mathop{{F}}\nolimits^{{G\mathop{{}}\nolimits_{{S}}}}}}}\nolimits_{{i}},i \neq 0}\\
{\mathop{{{\mathop{{F}}\nolimits^{{G\mathop{{}}\nolimits_{{S}}}}}}}\nolimits_{{i}},i=0}
\end{array}\right. }},\\
{\mathop{{{\mathop{{F}}\nolimits^{{B}}}}}\nolimits_{{i}}={ \left\{ \begin{array}{*{20}{l}}
{\mathop{{{\mathop{{F}}\nolimits^{{B}}}}}\nolimits_{{i-1}} \oplus \mathop{{{\mathop{{F}}\nolimits^{{S}}}}}\nolimits_{{i-1}} \oplus \mathop{{{\mathop{{F}}\nolimits^{{G\mathop{{}}\nolimits_{{B}}}}}}}\nolimits_{{i}},i \neq 0}\\
{\mathop{{{\mathop{{F}}\nolimits^{{G\mathop{{}}\nolimits_{{B}}}}}}}\nolimits_{{i}},i=0}
\end{array}\right. }},\\
{\mathop{{{\mathop{{F}}\nolimits^{{G}}}}}\nolimits_{{i}}=Cat \left( \mathop{{{\mathop{{F}}\nolimits^{{S}}}}}\nolimits_{{i}},\mathop{{{\mathop{{F}}\nolimits^{{B}}}}}\nolimits_{{i}} \right) },
\end{gather}
where the $\oplus$ is element-wise sum, and the $Cat$ is concatenation operation. $F^{G_S}$$_i$ is the guide saliency features generated by saliency feature stream, and $F^{G_B}$$_i$ is the guide boundary features generated by boundary feature stream.

In the first Guide Block, the saliency features $F^S$$_0$ and boundary features $F^B$$_0$ are directly set to equal to the $F^{G_S}$$_0$ and $F^{G_B}$$_0$. After that, we add the guide saliency features $F^{G_S}$$_i$ to the $F^S$$_{i-1}$ generated in the previous recurrent step, aiming to generate the more detailed saliency features $F^S$$_i$. Then we add $F^B$$_{i-1}$ to $F^S$$_i$ in order to supplement missing boundary information. On the other hand, we also add $F^S$$_{i-1}$ to $F^B$$_i$ to supplement missing saliency information, and use $F^{G_B}$$_i$ to generate more detailed boundary information.

Furthermore, the saliency features $F^S$$_i$ contain rich information of salient object, and boundary features $F^B$$_{i}$ contain more details of boundary. In order to sufficiently utilize these two different features, the guide features $F^{G}$$_i$ is then obtained by concatenating $F^S$$_i$ and $F^B$$_{i}$ together, used to guide the prediction of salient and boundary feature streams. Since $F^{G}$$_i$ include both information of salient object and boundary, they effectively guide the prediction of the feature stream of both saliency and boundary.

In addition, to obtain guide saliency features $F^{G_S}$$_i$ and guide boundary features $F^{G_B}$$_i$ which contain richer information, the concatenation ($Cat$) of low-level integrated features and the guide features $F^G$$_{i-1}$ at the previous step are fed to the saliency feature stream and the boundary feature stream. These feature streams consist of convolution operation, they can be defined as:
 
\begin{gather}
{{F\mathop{{}}\nolimits^{{G\mathop{{}}\nolimits_{{S}}}}\mathop{{}}\nolimits_{{i}}=}{ \left\{ \begin{array}{*{20}{l}}{ \Phi \mathop{{}}\nolimits^{{S}}\mathop{{}}\nolimits_{{i}} \left( Cat \left( F\mathop{{}}\nolimits^{{G}}\mathop{{}}\nolimits_{{i-1}},F\mathop{{}}\nolimits^{{L}} \left)  \left) ,i \neq 0\right. \right. \right. \right. }\\
{ \Phi \mathop{{}}\nolimits^{{S}}\mathop{{}}\nolimits_{{i}} \left( F\mathop{{}}\nolimits^{{H}} \left) ,i=0\right. \right. }
\end{array}\right. }},\\
{{F\mathop{{}}\nolimits^{{G\mathop{{}}\nolimits_{{B}}}}\mathop{{}}\nolimits_{{i}}=}{ \left\{ \begin{array}{*{20}{l}}
{ \Phi \mathop{{}}\nolimits^{{B}}\mathop{{}}\nolimits_{{i}} \left( Cat \left( F\mathop{{}}\nolimits^{{G}}\mathop{{}}\nolimits_{{i-1}},F\mathop{{}}\nolimits^{{L}} \left)  \left) ,i \neq 0\right. \right. \right. \right. }\\
{ \Phi \mathop{{}}\nolimits^{{B}}\mathop{{}}\nolimits_{{i}} \left( F\mathop{{}}\nolimits^{{H}} \left) ,i=0\right. \right. }
\end{array}\right. }},
\end{gather}
where $\Phi$$^S$$_i$ and $\Phi$$^B$$_i$ refer to the salient feature stream and boundary feature stream. $F^H$ is the high-level integrated features and $F^L$ is the low-level integrated features.

The first guide salient features $F^{G_S}$$_0$ and guide boundary features $F^{G_B}$$_0$ are generated by $F^H$. At each recurrent step, guide salient features $F^{G_S}$$_i$ and guide boundary features $F^{G_B}$$_i$ are obtained by sending the concatanation $Cat$ of the $F^L$ and guide features $F^G$$_{i-1}$. To better demonstrate the effectiveness of our TGRM, we will give some comparisons later.

\subsection{Recurrent Two-Stream Guided Refinement Network}

In order to refine saliency features and boundary features simultaneously, we then develop a novel network with a sequence of TGRMs.  For an input image $I$, we first use ResNeXt~\cite{xie2017aggregated} as the feature extraction network. Since the multi-level features extracted from feature extraction network contains various information of salient objects and boundaries, we use them for different tasks. 

The high-level integrated features formed by features extracted from deeper layers are used to generate the initial salient object and boundary features. These features include rich semantic information, thus can provide regional position of salient object and boundary. Meanwhile, since features from shallower layers contain more details information, these features are merged together to form the low-level integrated features. At each recurrent step, although ~\cite{deng18r} utilizes both low-level integrated features and high-level integrated features to refine the initial saliency map, the redundant high-level integrated features which contain too much semantic cues of the salient objects will weaken the prediction of pixels in complex scenes. In spired by this observation, we only utilize the low-level integrated features as a reference to get better details.

At each recurrent step, the salient object features and boundary features generated by previous TGRM are refined in the Guide Block. The first Guide Block takes the intial salient object features and boundary features generated by high-level integrated features as input, while other Guide Blocks utilize guided features generated by two feature streams to refine these initial features. As the number of recurrent steps increases, these features which are fed into the Guide Block contain more and more details and complementary information of another task, so that the details and complementary information can help the network to generate more accurate saliency map and more clear boundary map.

\subsection{Loss Function}
As shown in Fig. 3, each Guide Block at the recurrent step outputs a saliency map $S_i$ and a boundary map $B_i$, which can be defined as a two-stream map $M_i$ = ($S_i$ , $B_i$). our network is designed to  output two-stream maps, including the initial two-stream map $M_0$, and a sequence of refined two-stream maps ($M_1$, ... , $M_n$) after iterating TGRM $n$ times. Since the wide appication of the cross-entroy loss in pixel-level vision task, we adopt it to calculate the difference between the saliency map and the ground truth, the difference between the final boundary map and the ground truth, Respectively. The total loss $L$ of our network is defined as the summation of the loss on all predicted saliency maps and boundary maps:
\begin{gather}
{{L\mathop{{}}\nolimits^{{M}}\mathop{{}}\nolimits_{{i}}=L\mathop{{}}\nolimits^{{S}}\mathop{{}}\nolimits_{{i}}}+L\mathop{{}}\nolimits^{{B}}\mathop{{}}\nolimits_{{i}}} ,\\
{L= \omega \mathop{{}}\nolimits_{{0}}L\mathop{{}}\nolimits^{{M}}\mathop{{}}\nolimits_{{0}}+{\mathop{ \sum }\limits_{{i=1}}^{{n}}{ \omega \mathop{{}}\nolimits_{{i}}L\mathop{{}}\nolimits^{{M}}\mathop{{}}\nolimits_{{i}}}}} ,
\end{gather}
where $L^S$$_i$ and $L^B$$_i$ denote the loss of the prediction of salient object and the prediction of boundary at $i$-th recurrent step, $\omega$$_i$ refers to the weight of each recurrent step and we empirically set all the weights to 1. Meanwhile, $n$ refers to the number of recurrent steps employed in our network. We set the hyper-parameter $n$ as 4 by keeping the balance between the time consumption and the detection accuracy. 

\begin{table*}[htbp]
  \centering
\scalebox{0.9}{
    \begin{tabular}{c|c|c|c|c|c|c|c|c|c|c|c|c}
    \hline
    \multirow{2}{*}{Method} & \multicolumn{2}{c|}{ECSSD} & \multicolumn{2}{c|}{HKU-IS} & \multicolumn{2}{c|}{DUT-OMRON} & \multicolumn{2}{c|}{PASCAL-S} & \multicolumn{2}{c|}{SOD} & \multicolumn{2}{c}{DUTS-test} \\
\cline{2-13}          & \textit{F$_\beta$$\uparrow$} & MAE$\downarrow$   & \textit{F$_\beta$$\uparrow$} & MAE$\downarrow$   & \textit{F$_\beta$$\uparrow$} & MAE$\downarrow$   & \textit{F$_\beta$$\uparrow$} & MAE$\downarrow$   & \textit{F$_\beta$$\uparrow$} & MAE$\downarrow$   & \textit{F$_\beta$$\uparrow$} & MAE$\downarrow$ \\
\hline
    ELD~\cite{lee2016deep}   & 0.865 & 0.082 & 0.840 & 0.074 & 0.738 & 0.092 & 0.773 & 0.123 & 0.767 & 0.152 & 0.747 & 0.093 \\
    RFCN~\cite{wang2016saliency}  & 0.890  & 0.108 & 0.892 & 0.089 & 0.744 & 0.111 & 0.837 & 0.118 & 0.801 & 0.170  & 0.784 & 0.091 \\
    DSS~\cite{HouPami19Dss}   & 0.916 & 0.052 & 0.910  & 0.041 & 0.771 & 0.066 & 0.836 & 0.096 & 0.844 & 0.121 & 0.825 & 0.057 \\
    NLDF~\cite{luo2017non}  & 0.903 & 0.065 & 0.902 & 0.048 & 0.753 & 0.079 & 0.831 & 0.099 & 0.837 & 0.123 & 0.816 & 0.055 \\
    Amulet~\cite{zhang2017amulet} & 0.911 & 0.062 & 0.889 & 0.052 & 0.737 & 0.083 & 0.837 & 0.098 & 0.799 & 0.146 & 0.773 & 0.075 \\
    R3Net~\cite{deng18r} & 0.935 & 0.040  & 0.916 & 0.036 & 0.805 & 0.063 & 0.845 & 0.097 & 0.847 & 0.124 & 0.828 & 0.059 \\
    PAGR~\cite{zhang2018progressive}  & 0.927 & 0.061 & 0.918 & 0.048 & 0.771 & 0.071 & 0.856 & 0.093 & 0.838 & 0.144 & 0.855 & 0.056 \\
    BMPM~\cite{zhang2018bi}  & 0.928 & 0.044 & 0.920  & 0.038 & 0.774 & 0.064 & 0.862 & 0.074 & 0.851 & 0.106 & 0.85  & 0.049 \\
    PiCANet~\cite{liu2018picanet} & 0.940  & \textcolor[rgb]{ 0,  0,  1}{\textbf{0.035}} & 0.927 & 0.031 & 0.804 & 0.054 & 0.867 & 0.067 & 0.858 & 0.109 & 0.866 & 0.041 \\
    AFNet+~\cite{Feng_2019_CVPR} & 0.935 & 0.040  & 0.929 & 0.030  & 0.824 & 0.053 & 0.866 & 0.067 & 0.864 & 0.107 & 0.870  & 0.041 \\
    MLMSNet+~\cite{WuRunMin_2019_CVPR} & 0.929 & 0.041 & 0.926 & 0.031 & 0.800   & 0.058 & 0.858 & 0.068 & 0.866 & \textcolor[rgb]{ 0,  1,  0}{\textbf{0.101}} & 0.859 & 0.043 \\
    PAGENet+~\cite{Wang_2019_CVPR} & 0.93  & 0.040  & 0.921 & 0.031 & 0.793 & 0.058 & 0.85  & 0.073 & 0.841 & 0.107 & 0.841 & 0.048 \\
    BASNet+~\cite{Qin_2019_CVPR} & 0.941 & 0.037 & 0.932 & 0.029 & 0.813 & 0.055 & 0.854 & 0.074 & 0.852 & 0.112 & 0.862 & 0.046 \\
    poolNet+~\cite{Liu2019PoolSal} & \textcolor[rgb]{ 0,  1,  0}{\textbf{0.946}} & \textcolor[rgb]{ 0,  1,  0}{\textbf{0.034}} & \textcolor[rgb]{ 0,  0,  1}{\textbf{0.940}} & \textcolor[rgb]{ 0,  1,  0}{\textbf{0.025}} & \textcolor[rgb]{ 0,  0,  1}{\textbf{0.835}} & \textcolor[rgb]{ 0,  0,  1}{\textbf{0.050}} & \textcolor[rgb]{ 0,  1,  0}{\textbf{0.882}} & \textcolor[rgb]{ 0,  1,  0}{\textbf{0.061}} & \textcolor[rgb]{ 0,  0,  1}{\textbf{0.879}} & 0.102 & \textcolor[rgb]{ 0,  1,  0}{\textbf{0.895}} & \textcolor[rgb]{ 0,  1,  0}{\textbf{0.033}} \\
    EGNet+~\cite{zhao2019EGNet} & \textcolor[rgb]{ 0,  0,  1}{\textbf{0.944}} & 0.037 & \textcolor[rgb]{ 0,  1,  0}{\textbf{0.939}} & \textcolor[rgb]{ 0,  0,  1}{\textbf{0.026}} & \textcolor[rgb]{ 0,  1,  0}{\textbf{0.843}} & \textcolor[rgb]{ 0,  1,  0}{\textbf{0.049}} & 0.869 & 0.07  & \textcolor[rgb]{ 0,  1,  0}{\textbf{0.893}} & \textcolor[rgb]{ 0,  0,  1}{\textbf{0.096}} & \textcolor[rgb]{ 0,  0,  1}{\textbf{0.893}} & 0.035 \\
    SCRN+~\cite{Wu_2019_ICCV} & \textcolor[rgb]{ 0,  1,  0}{\textbf{0.946}} & \textcolor[rgb]{ 0,  0,  1}{\textbf{0.035}} & \textcolor[rgb]{ 0,  1,  0}{\textbf{0.939}} & \textcolor[rgb]{ 0,  0,  1}{\textbf{0.026}} & 0.815 & \textcolor[rgb]{ 0,  1,  0}{\textbf{0.049}} & \textcolor[rgb]{ 0,  0,  1}{\textbf{0.879}} & \textcolor[rgb]{ 1,  0,  0}{\textbf{0.058}} & 0.87  & \textcolor[rgb]{ 0,  1,  0}{\textbf{0.101}} & 0.890  & \textcolor[rgb]{ 0,  0,  1}{\textbf{0.034}} \\
\hline
    RTGRNet & \textcolor[rgb]{ 1,  0,  0}{\textbf{0.950}} & \textcolor[rgb]{ 1,  0,  0}{\textbf{0.031}} & \textcolor[rgb]{ 1,  0,  0}{\textbf{0.944}} & \textcolor[rgb]{ 1,  0,  0}{\textbf{0.024}} & \textcolor[rgb]{ 1,  0,  0}{\textbf{0.848}} & \textcolor[rgb]{ 1,  0,  0}{\textbf{0.048}} & \textcolor[rgb]{ 1,  0,  0}{\textbf{0.885}} & \textcolor[rgb]{ 0,  0,  1}{\textbf{0.066}} & \textcolor[rgb]{ 1,  0,  0}{\textbf{0.894}} & \textcolor[rgb]{ 1,  0,  0}{\textbf{0.091}} & \textcolor[rgb]{ 1,  0,  0}{\textbf{0.898}} & \textcolor[rgb]{ 1,  0,  0}{\textbf{0.032}} \\
    \hline
    \end{tabular}}%
  \label{tab:addlabel}%
\caption{Comparisons to the state-of-the-art methods on six public datasets for salient object detection. The top three results are highlighted in \textcolor[rgb]{ 1,  0,  0}{\textbf{red}}, \textcolor[rgb]{ 0,  1,  0}{\textbf{green}}, and \textcolor[rgb]{ 0,  0,  1}{\textbf{blue}}. The same post-processing method CRF is used in AFNet, MLMSNet,PAGENet, BASNet, poolNet, EGNet and SCRN.}
\end{table*}%

\section{Experiments}
In this section, we first describe the experiment setups of our proposed network, including the implementation details, the used datasets and the evaluation metric. After that, we compare our network with the previous state-of-the-art methods in order to show the effect of our network. At last,we will conduct a series of ablation studies to validate the effectiveness of each component of our proposed network on the salient object detection task.
\subsection{Experiment Setup}
{\bf Implementation Details.} The parameters of our backbone network are initialized with the ResNeXt~\cite{xie2017aggregated} network pretrained on ImageNet, and parameters of other layers are initialized by Gaussian distribution. Before training, images in training set are resized to $300 \times 300$, and randomly cropped, rotated, horizontally flipped for data augmentation. Our network is built on the public platform Pytorch and the hyper-parameters are set as follows: a base learning rate = 0.001, momentum parameter = 0.9, weight decay = 0.001, and batchsize = 12. Our RTGRNet is trained on a single GTX 1080Ti GPU (with 11GB memory) for 15$\sim$20K iterations, and takes 4 hours to train the network. The SGD method is used to train our network. When testing, each image is resized to $300 \times 300$ before fed into our network.

{\bf Dataset.} To evaluate the performance of our proposed network, we conduct evaluations on six public salient object benchmark dateset: ECSSD~\cite{yan2013hierarchical}, HKU-IS~\cite{li2015visual}, DUT-OMRON~\cite{yang2013saliency}, SOD~\cite{movahedi2010design}, PASCAL-S~\cite{li2014secrets} and DUTS~\cite{wang2017}. The first five datasets are widely used in salient object detection. DUTS is currently the largest saliency detection dataset, composed of DUTS-train and DUTS-test. DUTS-train contains 10553 images and DUTS-test contains 5019 imgaes. Since many deep salient object methods are trained on the training set of DUTS, we also train our model on the training set of DUTS and use its testing set as one of our testing datasets. Specifically, we evalute the performance using the following metrics.

{\bf Evaluation Metrics.} To evaluate our network, three popular criteria are used for performance evalution, i.e. F-measure, mean absolute error (MAE) and precision and recall curve (short for PR curve). F-measure, denoted by F$_\beta$, is an overall performance measurement and comprehensively considers both Precision and Recall by computing the weighted harmonic mean:
\begin{gather}
{F\mathop{{}}\nolimits_{{ \beta }}=\frac{{ \left( 1+ \beta \mathop{{}}\nolimits^{{2}} \left)  \times Precision \times Recall\right. \right. }}{{ \beta \mathop{{}}\nolimits^{{2}} \times Precision+Recall}}},
\end{gather}

where $\beta$$^2$ is set to 0.3, consistent with the usual settings for this parameter. Mean absolute error (MAE) is computed by:
\begin{gather}
{MAE=\frac{{1}}{{W \times H}}{\mathop{ \sum }\limits_{{x=1}}^{{W}}{{\mathop{ \sum }\limits_{{y=1}}^{{H}}{{ \left| {P \left( x,y \left) -Y \left( x,y \right) \right. \right. } \right| }}}}}},
\end{gather}

where Y is the ground truth (GT), and P is the saliency map of network output. Finally, Precision-Recall (PR) is a standard metric to evlauate saliency performance, calculated based on the binarized saliency map and the ground-truth:
\begin{gather}
{Precision=\frac{{TP}}{{TP+FP}} , Recall=\frac{{TP}}{{TP+FN}}} ,
\end{gather}
where TP, TN, FP, FN denote true-positive, true-negative, false-positive, and false-negetive, respectively.

{\bf Inference.} Given an image as input, our RTGRNet predicts a saliency map and a boundary map at each recurrent step, and the saliency map producted by the last recurrent step is taken as our final result. the fully connected conditional random field (CRF)~\cite{krahenbuhl2011efficient} is then used to enhance the spatial coherence of the saliency maps.
\begin{figure*}
\begin{center}
\includegraphics[width=1\textwidth]{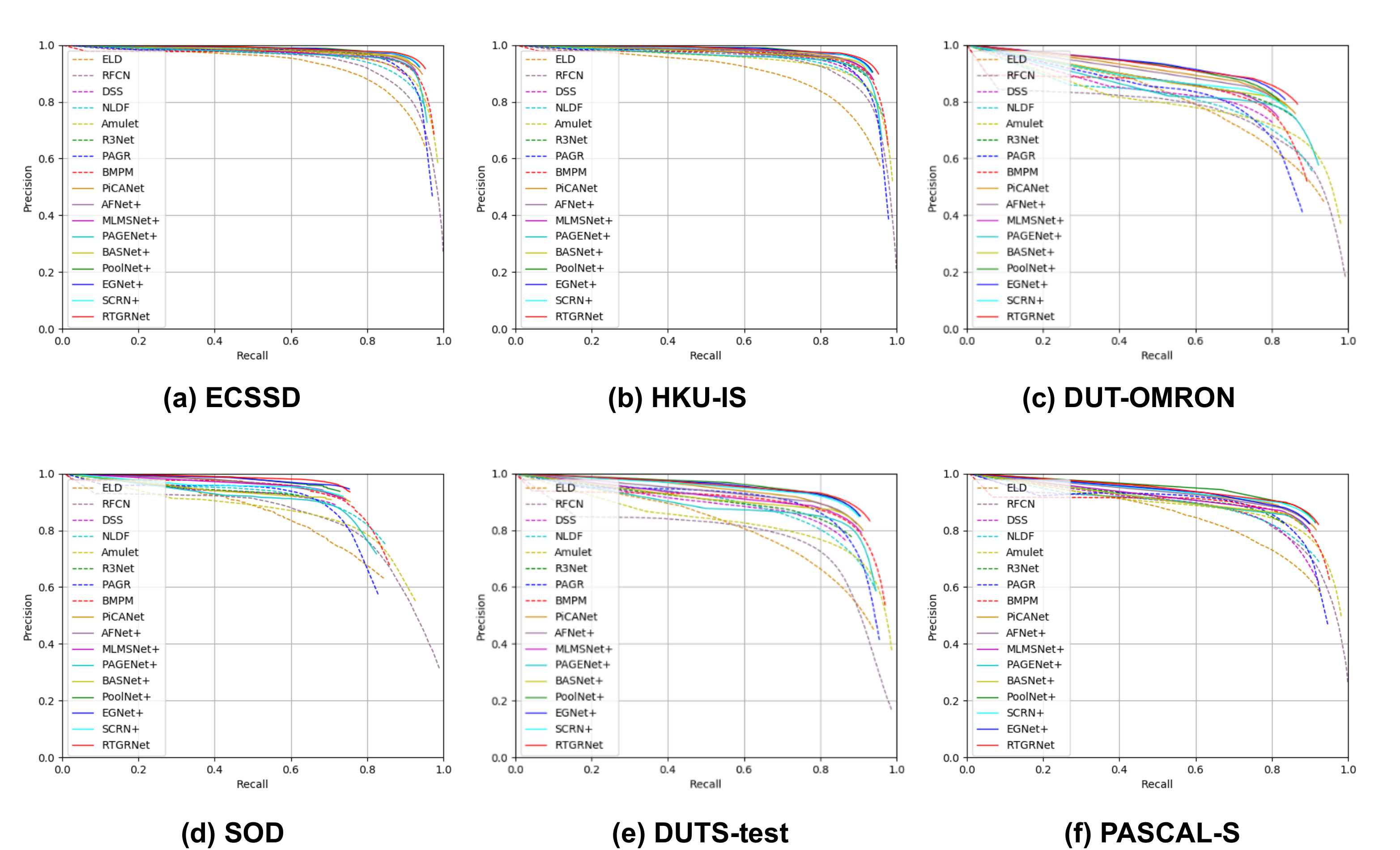} 
\end{center}
   \caption{The PR Curves of our RTGRNet and other state-of-the-art methods over six dataset.}
\label{fig:short}
\end{figure*}
\begin{figure*}
\begin{center}
\includegraphics[width=0.9\textwidth]{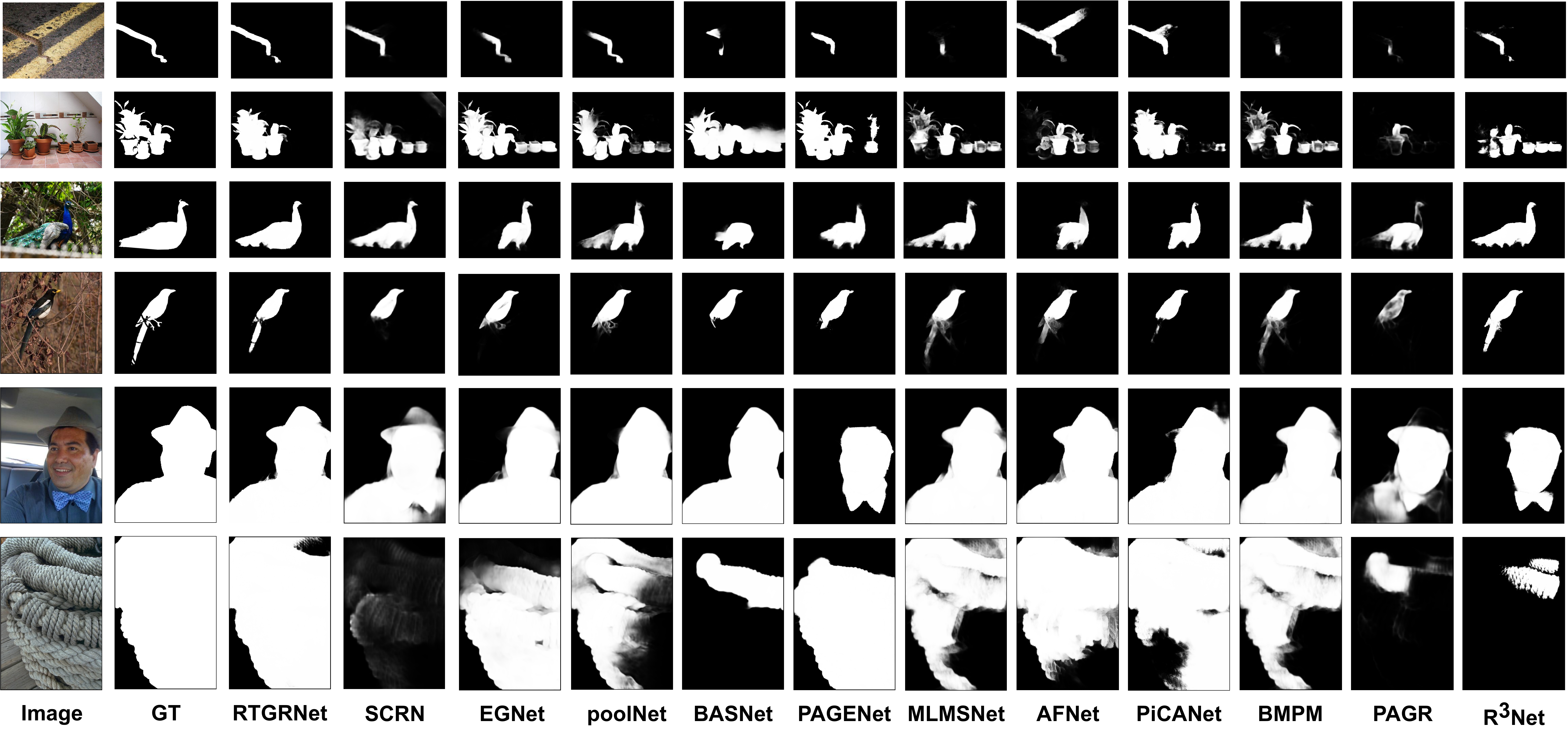} 
\end{center}
   \caption{Visual comparisons of our RTGRNet and the state-of-the-art methods.}
\label{fig:short}
\end{figure*}
\subsection{Comparisons with State-of-the-Art Methods}
We first compare the results of our method with other 16 state-of-the-art salient object detection networks, which are ELD~\cite{lee2016deep}, RFCN~\cite{wang2016saliency}, DSS~\cite{HouPami19Dss}, NLDF~\cite{luo2017non}, Amulet~\cite{zhang2017amulet}, R$^3$Net~\cite{deng18r}, PAGR~\cite{zhang2018progressive}, BMPM~\cite{zhang2018bi}, PiCANet~\cite{liu2018picanet}, AFNet~\cite{Feng_2019_CVPR}, MLMSNet~\cite{WuRunMin_2019_CVPR}, PAGENet~\cite{Wang_2019_CVPR}, BASNet~\cite{Qin_2019_CVPR}, poolNet~\cite{Liu2019PoolSal}, EGNet~\cite{zhao2019EGNet} and SCRN~\cite{Wu_2019_ICCV}. All the saliency maps are provided by the authors.

{\bf Quantitative Evaluation.} To evaluate the quality of detected salient objects, we show the PR curves in Fig. 4. Meanwhile, the comparisons between our RTGRNet and 16 current state-of-the-art methods in terms of F-measure and MAE score are shown in Table 1. For fair comparisons, all methods proposed in 2019 (shown as AFNet+, MLMSNet+, PAGENet+, BASNet+, poolNet+, EGNet+ and SCRN+) apply the same post-processing denseCRF to refine final saliency maps. It is worth mentioning that MLMSNet and poolNet need extra edge detection dataset to train another boundary detection branch. In addition, EGNet and SCRN focus on the interaction between information of salient object and boundary. It is evident that our RTGRNet achieves excellent results for all the datasets, and shows a significantly improvement in both F$_\beta$ and MAE  when compared to the second best method. Besides, all the Evaluation Metric demonstrates that our RTGRNet has superior performance in different challenging scenes.
\begin{table}[htbp]
  \centering
    \begin{tabular}{|c|c|c|c|c|}
    \hline
    \multirow{2}{*}{Module} & \multicolumn{2}{c|}{DUT-OMRON} & \multicolumn{2}{c|}{DUTS-test} \\
\cline{2-5}          & \textit{F$_\beta$$\uparrow$} & MAE$\downarrow$   & \textit{F$_\beta$$\uparrow$} & MAE$\downarrow$ \\
    \hline
    RRB   & 0.836 & 0.053 & 0.891 & 0.036 \\
    SGRM  & 0.839 & 0.051 & 0.895  & 0.034 \\
    TGRM  & \textbf{0.848} & \textbf{0.048} & \textbf{0.898} & \textbf{0.032} \\
    \hline
    \end{tabular}%
  \label{tab:addlabel}%
 \caption{Ablation analysis for different recurrent module on two popular datasets.}
\end{table}%

\begin{table*}[htbp]
  \centering
    \begin{tabular}{c|c|c|c|c|c|c|c|c|c|c}
    \hline
    \multirow{2}{*}{Method} & \multicolumn{2}{c|}{ECSSD} & \multicolumn{2}{c|}{HKU-IS} & \multicolumn{2}{c|}{DUT-OMRON} & \multicolumn{2}{c|}{PASCAL-S} & \multicolumn{2}{c}{DUTS-test} \\
\cline{2-11}          & \textit{F$_\beta$$\uparrow$} & MAE$\downarrow$  & \textit{F$_\beta$$\uparrow$} & MAE$\downarrow$   & \textit{F$_\beta$$\uparrow$} & MAE$\downarrow$   & \textit{F$_\beta$$\uparrow$} & MAE$\downarrow$   & \textit{F$_\beta$$\uparrow$} & MAE$\downarrow$ \\
    \hline
    RTGRNet-0 & 0.936 & 0.037 & 0.930  & 0.030  & 0.830  & 0.053 & 0.871 & 0.095 & 0.880  & 0.037 \\
  RTGRNet-1 & 0.945 & 0.034 & 0.940  & 0.026 & 0.842 & 0.049 & 0.882 & 0.059 & 0.893 & 0.034 \\
RTGRNet-2 & 0.947 & 0.032 & 0.942 & 0.025 & 0.842 & 0.049 & 0.883 & 0.06 & 0.892 & 0.035 \\
 RTGRNet-3 & 0.948 & \textbf{0.031} & \textbf{0.944} & \textbf{0.024} & 0.844 & 0.049 & 0.883 & \textbf{0.057} & 0.892 & 0.035 \\
RTGRNet-4 & \textbf{0.950} & \textbf{0.031} & \textbf{0.944} & \textbf{0.024} & \textbf{0.848} & \textbf{0.048} & \textbf{0.885} & 0.066 & \textbf{0.898} & \textbf{0.032} \\
 RTGRNet-5 & 0.949 & \textbf{0.031} & 0.943 & 0.025 & 0.846 & 0.051 & 0.884 & 0.061 & 0.894 & 0.035 \\
    \hline
    \hline
    RTGRNet-hh & 0.937 & 0.038 & 0.929 & 0.031 & 0.834 & 0.051 & 0.876 & 0.063 & 0.882 & 0.037 \\
    RTGRNet-ll & 0.875 & 0.076 & 0.881 & 0.053 & 0.713 & 0.099 & 0.792 & 0.137 & 0.781 & 0.075 \\
    RTGRNet-hl-2 & 0.947 & 0.032 & 0.943 & 0.025 & 0.841 & 0.05  & 0.88  & 0.06 & 0.892 & 0.034 \\
    \hline
    \end{tabular}%
  \label{tab:addlabel}%
\caption{Ablation experiment of RTGRNet on six datasets.}
\end{table*}%

{\bf Visual Comparisons.} To further illustrate the performance of our method, Fig. 5 shows the visual comparisons of our RTGRNet with other top twelve state-of-the-art methods. The images correspond to scenes with low contrast between object and background (the first, fifth and the last row), complex background (the third row and the fourth row), big object (the fifth row and the last row) and multiple objects (the second row). It can be seen that our methods can highlight salient objects and keep clear boundaries, especially in scenes with low contrast between foreground and background.

\subsection{Ablation Studies}
In this section, we analyze each module in the network to prove their validity.

{\bf Effectiveness of Boundary Feature Stream.} To validate the effectiveness of our boundary feature stream, we compare three variants of recurrent module: RRB, SGRM and TGRM. Baseline RRB uses saliency map predicted by previous recurrent step to guide the prediction of next step. Based on the RRB, SGRM uses a 32-channel feature map to replace the saliency map to guide the prediction of next step. For TGRM, we add a boundary feature stream based on SGRM to enable boundary features and salient features to guide each other. As shown in Table 2, the network with TGRM achieves better performance compared with those with RRB or SGRM. This result demonstrates that boundary feature stream can boost performance effectively. Meanwhile, we show some examples of our generated saliency map and boundary map in Fig. 6. As can be seen, our network generates clear boundary and achieves superior detection result. 

{\bf Effectiveness of Recurrent mechanism.} We perform ablation experiments on the six public datasets to evaluate the effectiveness of recurrent mechanism. From the results shown in Table 2, we have the following observations. (1) First, from the first row to the sixth row, we evaluate the effectiveness of each recurrent step. Among them, RTGRNet-0 means that we only use the initial saliency map generated by the high-level integrated features as our final result, and RTGRNet-i (i = 1, 2, 3, ..., n) means that our network contains $i$ TGRM modules guided by low-level integrated features and the guide features generated by the Guided block. It can be observed that as the number of TGRM increases, the performance of RTGRNet on each dataset is gradually improved, and then becomes stable from the fourth or 
\begin{figure}
\begin{center}
\includegraphics[width=0.49\textwidth]{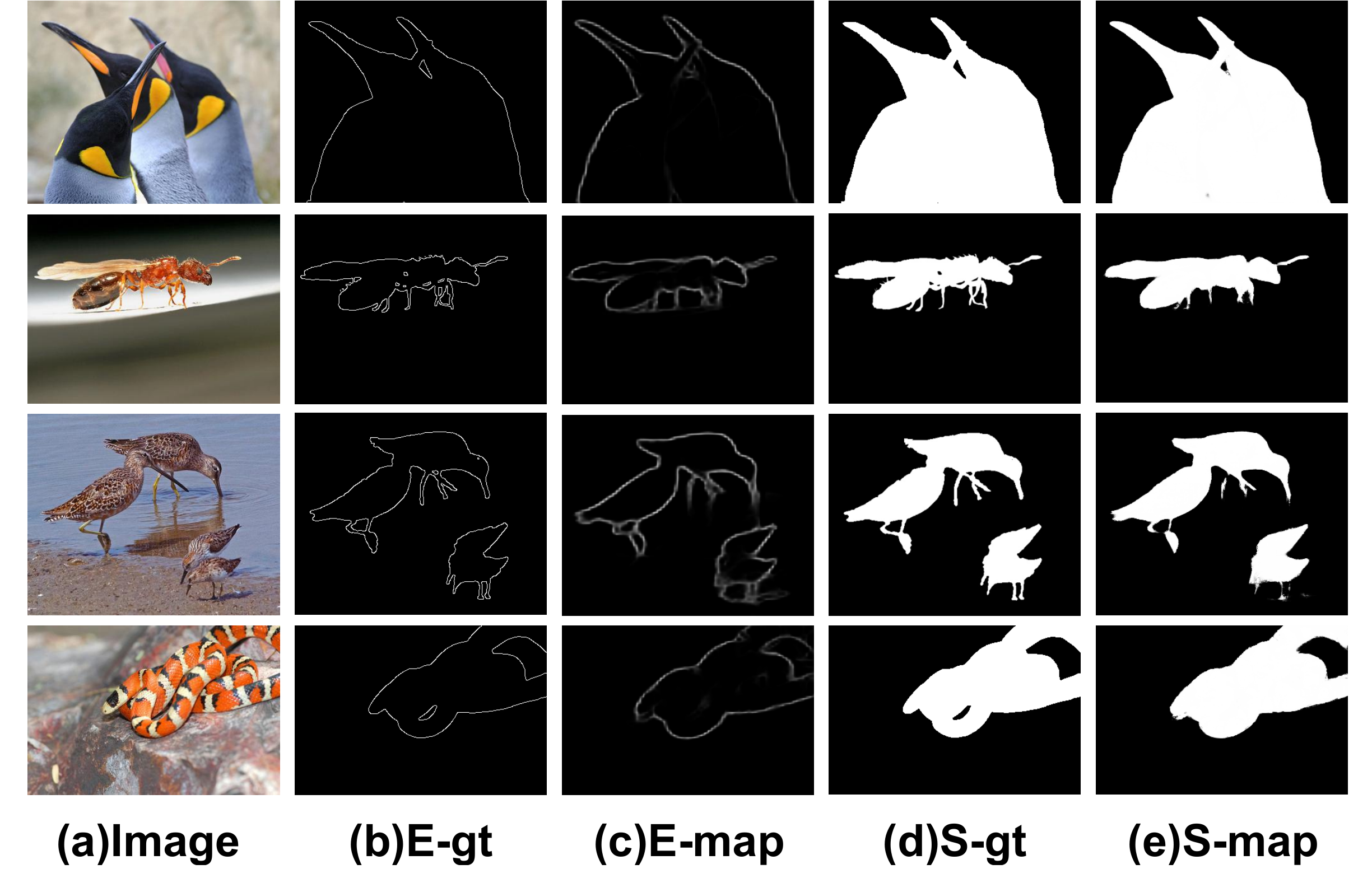} 
\end{center}
   \caption{Examples of our generated saliency map and boundary map. Image and S-gt are collected from salient object detection datasets and E-gt is the salient object's boundary extracted from S-gt. E-map and S-map are boundary map and saliency map generated by our model.}
\label{fig:short}
\end{figure}
the fifth recurrent step, which proves that our recurrent mechanism can effectively improve the detection capability of RTGRNet. (2) After that, we also compare our RTGRNet with three different recurrent strategies RTGRNet-HH (using high-level integrated features to refine the initial saliency map), RTGRNet-LL (using low-level integrated features to produce the initial saliency map) and RTGRNet-HL-2 (low-level and high-level integrated features are iteratively used to refine the initial saliency map).As we can see,  Compare with RTGRNet-HH and RTGRNet-LL, RTGRNet-1 achieves the best results. Meanwhile, compare with RTGRNet-HL-2, our RTGRNet-3 which executes the same recurrent step achieves the best results.

\section{Conclusion}
In this paper, we present a novel boundary-aware method, namely RTGRNet, for salient object detection. RTGRNet is equipped with a sequence of two-stream guided refinement module (TGRM), each TGRM consists of a Guide Block, a saliency feature stream and a boundary feature stream. These two different feature streams are utilized to guide each other to complement the lack of boundary information and salient object information, respectively. Furthermore, we use TGRMs to recurrently refine the initial saliency map and boundary map generated by high-level features. In this way, RTGRNet is able to output high quality saliency maps and boundary maps. Finally, experimental results on six public datasets demonstrate that our method outperforms the other 16 state-of-the-art methods.
\newpage
{\small
\bibliographystyle{ieee_fullname}
\bibliography{references}

\begin{thebibliography}{10}\itemsep=-1pt

\bibitem{borji2014salient}
Ali Borji, Ming-Ming Cheng, Qibin Hou, Huaizu Jiang, and Jia Li.
\newblock Salient object detection: A survey.
\newblock {\em Computational Visual Media}, 2014.

\bibitem{cao2017realtime}
Zhe Cao, Tomas Simon, Shih-En Wei, and Yaser Sheikh.
\newblock Realtime multi-person 2d pose estimation using part affinity fields.
\newblock In {\em CVPR}, pages 7291--7299, 2017.

\bibitem{cheng2014global}
Ming-Ming Cheng, Niloy~J Mitra, Xiaolei Huang, Philip~HS Torr, and Shi-Min Hu.
\newblock Global contrast based salient region detection.
\newblock {\em IEEE TPAMI}, pages 569--582, 2014.

\bibitem{deng18r}
Zijun Deng, Xiaowei Hu, Lei Zhu, Xuemiao Xu, Jing Qin, Guoqiang Han, and
  Pheng-Ann Heng.
\newblock R$^{3}${N}et: Recurrent residual refinement network for saliency
  detection.
\newblock In {\em IJCAI}, 2018.

\bibitem{fan2019shifting}
Deng-Ping Fan, Wenguan Wang, Ming-Ming Cheng, and Jianbing Shen.
\newblock Shifting more attention to video salient object detection.
\newblock In {\em Proceedings of the IEEE Conference on Computer Vision and
  Pattern Recognition}, pages 8554--8564, 2019.

\bibitem{Feng_2019_CVPR}
Mengyang Feng, Huchuan Lu, and Errui Ding.
\newblock Attentive feedback network for boundary-aware salient object
  detection.
\newblock In {\em CVPR}, 2019.

\bibitem{HouPami19Dss}
Qibin Hou, Ming-Ming Cheng, Xiaowei Hu, Ali Borji, Zhuowen Tu, and Philip Torr.
\newblock Deeply supervised salient object detection with short connections.
\newblock pages 815--828, 2019.

\bibitem{itti1998model}
Laurent Itti, Christof Koch, and Ernst Niebur.
\newblock A model of saliency-based visual attention for rapid scene analysis.
\newblock {\em IEEE TPAMI}, pages 1254--1259, 1998.

\bibitem{krahenbuhl2011efficient}
Philipp Kr{\"a}henb{\"u}hl and Vladlen Koltun.
\newblock Efficient inference in fully connected crfs with gaussian edge
  potentials.
\newblock In {\em Advances in neural information processing systems}, pages
  109--117, 2011.

\bibitem{lee2016deep}
Gayoung Lee, Yu-Wing Tai, and Junmo Kim.
\newblock Deep saliency with encoded low level distance map and high level
  features.
\newblock In {\em CVPR}, pages 660--668, 2016.

\bibitem{li2015visual}
Guanbin Li and Yizhou Yu.
\newblock Visual saliency based on multiscale deep features.
\newblock In {\em CVPR}, pages 5455--5463, 2015.

\bibitem{li2014secrets}
Yin Li, Xiaodi Hou, Christof Koch, James~M Rehg, and Alan~L Yuille.
\newblock The secrets of salient object segmentation.
\newblock In {\em Proceedings of the IEEE Conference on Computer Vision and
  Pattern Recognition}, pages 280--287, 2014.

\bibitem{liu2019coherent}
Hongyu Liu, Bin Jiang, Yi Xiao, and Chao Yang.
\newblock Coherent semantic attention for image inpainting.
\newblock {\em ICCV}, 2019.

\bibitem{Liu2019PoolSal}
Jiang-Jiang Liu, Qibin Hou, Ming-Ming Cheng, Jiashi Feng, and Jianmin Jiang.
\newblock A simple pooling-based design for real-time salient object detection.
\newblock In {\em CVPR}, 2019.

\bibitem{liu2016dhsnet}
Nian Liu and Junwei Han.
\newblock Dhsnet: Deep hierarchical saliency network for salient object
  detection.
\newblock In {\em CVPR}, pages 678--686, 2016.

\bibitem{liu2018picanet}
Nian Liu, Junwei Han, and Ming-Hsuan Yang.
\newblock Picanet: Learning pixel-wise contextual attention for saliency
  detection.
\newblock In {\em CVPR}, pages 3089--3098, 2018.

\bibitem{long2015fully}
Jonathan Long, Evan Shelhamer, and Trevor Darrell.
\newblock Fully convolutional networks for semantic segmentation.
\newblock In {\em CVPR}, pages 3431--3440, 2015.

\bibitem{luo2017non}
Zhiming Luo, Akshaya~Kumar Mishra, Andrew Achkar, Justin~A Eichel, Shaozi Li,
  and Pierre-Marc Jodoin.
\newblock Non-local deep features for salient object detection.
\newblock In {\em CVPR}, page~7, 2017.

\bibitem{movahedi2010design}
Vida Movahedi and James~H Elder.
\newblock Design and perceptual validation of performance measures for salient
  object segmentation.
\newblock In {\em CVPR}, pages 49--56, 2010.

\bibitem{Qin_2019_CVPR}
Xuebin Qin, Zichen Zhang, Chenyang Huang, Chao Gao, Masood Dehghan, and Martin
  Jagersand.
\newblock Basnet: Boundary-aware salient object detection.
\newblock In {\em CVPR}, 2019.

\bibitem{su2019selectivity}
Jinming Su, Jia Li, Yu Zhang, Changqun Xia, and Yonghong Tian.
\newblock Selectivity or invariance: Boundary-aware salient object detection.
\newblock In {\em Proceedings of the IEEE International Conference on Computer
  Vision}, 2019.

\bibitem{wang2017}
Lijun Wang, Huchuan Lu, Yifan Wang, Mengyang Feng, Dong Wang, Baocai Yin, and
  Xiang Ruan.
\newblock Learning to detect salient objects with image-level supervision.
\newblock In {\em CVPR}, 2017.

\bibitem{wang2016saliency}
Linzhao Wang, Lijun Wang, Huchuan Lu, Pingping Zhang, and Xiang Ruan.
\newblock Saliency detection with recurrent fully convolutional networks.
\newblock In {\em ECCV}, pages 825--841, 2016.

\bibitem{wang2018salient}
Linzhao Wang, Lijun Wang, Huchuan Lu, Pingping Zhang, and Xiang Ruan.
\newblock Salient object detection with recurrent fully convolutional networks.
\newblock {\em IEEE transactions on pattern analysis and machine intelligence},
  2018.

\bibitem{wang2011image}
Meng Wang, Janusz Konrad, Prakash Ishwar, Kevin Jing, and Henry Rowley.
\newblock Image saliency: From intrinsic to extrinsic context.
\newblock In {\em CVPR}, pages 417--424, 2011.

\bibitem{wang2017stagewise}
Tiantian Wang, Ali Borji, Lihe Zhang, Pingping Zhang, and Huchuan Lu.
\newblock A stagewise refinement model for detecting salient objects in images.
\newblock In {\em ICCV}, pages 4019--4028, 2017.

\bibitem{wang2017video}
Wenguan Wang, Jianbing Shen, and Ling Shao.
\newblock Video salient object detection via fully convolutional networks.
\newblock {\em IEEE Transactions on Image Processing}, 27(1):38--49, 2017.

\bibitem{Wang_2019_CVPR}
Wenguan Wang, Shuyang Zhao, Jianbing Shen, Steven C.~H. Hoi, and Ali Borji.
\newblock Salient object detection with pyramid attention and salient edges.
\newblock In {\em CVPR}, 2019.

\bibitem{wang2019focal}
Yupei Wang, Xin Zhao, Xuecai Hu, Yin Li, and Kaiqi Huang.
\newblock Focal boundary guided salient object detection.
\newblock {\em IEEE TIP}, pages 2813--2824, 2019.

\bibitem{WuRunMin_2019_CVPR}
Runmin Wu, Mengyang Feng, Wenlong Guan, Dong Wang, Huchuan Lu, and Errui Ding.
\newblock A mutual learning method for salient object detection with
  intertwined multi-supervision.
\newblock In {\em CVPR}, 2019.

\bibitem{Wu_2019_ICCV}
Zhe Wu, Li Su, and Qingming Huang.
\newblock Stacked cross refinement network for edge-aware salient object
  detection.
\newblock In {\em The IEEE International Conference on Computer Vision (ICCV)},
  2019.

\bibitem{xie2017aggregated}
Saining Xie, Ross Girshick, Piotr Doll{\'a}r, Zhuowen Tu, and Kaiming He.
\newblock Aggregated residual transformations for deep neural networks.
\newblock In {\em CVPR}, pages 1492--1500, 2017.

\bibitem{xie15hed}
Saining Xie and Zhuowen Tu.
\newblock Holistically-nested edge detection.
\newblock In {\em ICCV}, 2015.

\bibitem{yan2013hierarchical}
Qiong Yan, Li Xu, Jianping Shi, and Jiaya Jia.
\newblock Hierarchical saliency detection.
\newblock In {\em CVPR}, pages 1155--1162, 2013.

\bibitem{yang2013saliency}
Chuan Yang, Lihe Zhang, Huchuan Lu, Xiang Ruan, and Ming-Hsuan Yang.
\newblock Saliency detection via graph-based manifold ranking.
\newblock In {\em CVPR}, pages 3166--3173, 2013.

\bibitem{zhang2017deep}
Jing Zhang, Yuchao Dai, and Fatih Porikli.
\newblock Deep salient object detection by integrating multi-level cues.
\newblock In {\em WACV}, pages 1--10, 2017.

\bibitem{zhang2018bi}
Lu Zhang, Ju Dai, Huchuan Lu, You He, and Gang Wang.
\newblock A bi-directional message passing model for salient object detection.
\newblock In {\em CVPR}, pages 1741--1750, 2018.

\bibitem{zhang2017amulet}
Pingping Zhang, Dong Wang, Huchuan Lu, Hongyu Wang, and Xiang Ruan.
\newblock Amulet: Aggregating multi-level convolutional features for salient
  object detection.
\newblock In {\em ICCV}, pages 202--211, 2017.

\bibitem{zhang2018progressive}
Xiaoning Zhang, Tiantian Wang, Jinqing Qi, Huchuan Lu, and Gang Wang.
\newblock Progressive attention guided recurrent network for salient object
  detection.
\newblock In {\em CVPR}, pages 714--722, 2018.

\bibitem{zhao2019EGNet}
Jia-Xing Zhao, Jiang-Jiang Liu, Deng-Ping Fan, Yang Cao, Jufeng Yang, and
  Ming-Ming Cheng.
\newblock Egnet:edge guidance network for salient object detection.
\newblock In {\em The IEEE International Conference on Computer Vision (ICCV)},
  2019.

\bibitem{zhao2019pyramid}
Ting Zhao and Xiangqian Wu.
\newblock Pyramid feature attention network for saliency detection.
\newblock In {\em Proceedings of the IEEE Conference on Computer Vision and
  Pattern Recognition}, pages 3085--3094, 2019.

\end{thebibliography}
}

\end{document}